\documentclass[11pt,a4paper]{article}
\usepackage[hyperref]{acl2021}
\usepackage{gb4e}
\noautomath
\usepackage{paralist}
\usepackage{mathtools}
\usepackage[inline]{enumitem}
\usepackage{times}
\usepackage{latexsym}
\usepackage{amsmath}
\usepackage[capitalize]{cleveref}
\usepackage{amssymb}
\usepackage{multirow,multicol}
\usepackage{graphicx}

\usepackage{array}
\usepackage{bm}
\usepackage{soul, color}
\usepackage{booktabs}
\usepackage[colorinlistoftodos]{todonotes}
\usepackage{enumitem}
\usepackage{amsthm}
\usepackage{caption}
\usepackage{subcaption}
\usepackage{tipa}
\usepackage{stackengine}
\usepackage{rotating}
\usepackage{tabularx}
\usepackage{microtype}

\newcommand\code[1]{\texttt{#1}}
\crefformat{section}{\S#2#1#3} %
\crefformat{subsection}{\S#2#1#3}
\crefformat{subsubsection}{\S#2#1#3}
\newcommand\dataset{CIDER}

\title{Knowledge Extraction from Dyadic Dialogues}
\title{Explicit and Implicit Knowledge Extraction from Dyadic Dialogues} %
\title{Leveraging Commonsense to Extract Implicit Knowledge
from Dialogues}
\title{DICE: Dialogue-level Inference using Commonsense-based Explanation}
\title{CRaDLE: Commonsense Reasoning for Dialogue-Level Explanation}
\title{CIDER: Commonsense Inference for\\ Dialogue Explanation and Reasoning}
\title{CONSIDER: CONtextualized commonSense Inference for Dialogue Explanation and Reasoning}
\title{CONTENDER: CONTextualized commonsensE Inference for Dialogue Explanation and Reasoning}
\title{Dialogue Explanation and Reasoning with \\ Contextualized Commonsense Inference}
\title{Contextual-commonsense Inference for Dialogue Explanation and Reasoning}
\title{CIDER: Commonsense~Inference for Dialogue~Explanation~and~Reasoning}
\author{Deepanway Ghosal$^\dagger$, Pengfei Hong$^\dagger$, Siqi Shen$^\triangle$,\\
\textbf{Navonil Majumder$^\dagger$,
  Rada Mihalcea$^\triangle$,
  Soujanya Poria$^\dagger$}\\[1ex] 
  $^\dagger$ Singapore University of Technology and Design, Singapore\\
  $^\triangle$ University of Michigan, USA\\
  \texttt{\{deepanway\_ghosal, pengfei\_hong\}@mymail.sutd.edu.sg}\\
  \texttt{\{navonil\_majumder, sporia\}@sutd.edu.sg}\\ \texttt{\{shensq,mihalcea\}@umich.edu} 
  }

\date{}

\aclfinalcopy

\begin{document}

\twocolumn[{%
\renewcommand\twocolumn[1][]{#1}%
\begin{center}
\maketitle
    \centering
     \includegraphics[width=0.95\linewidth]{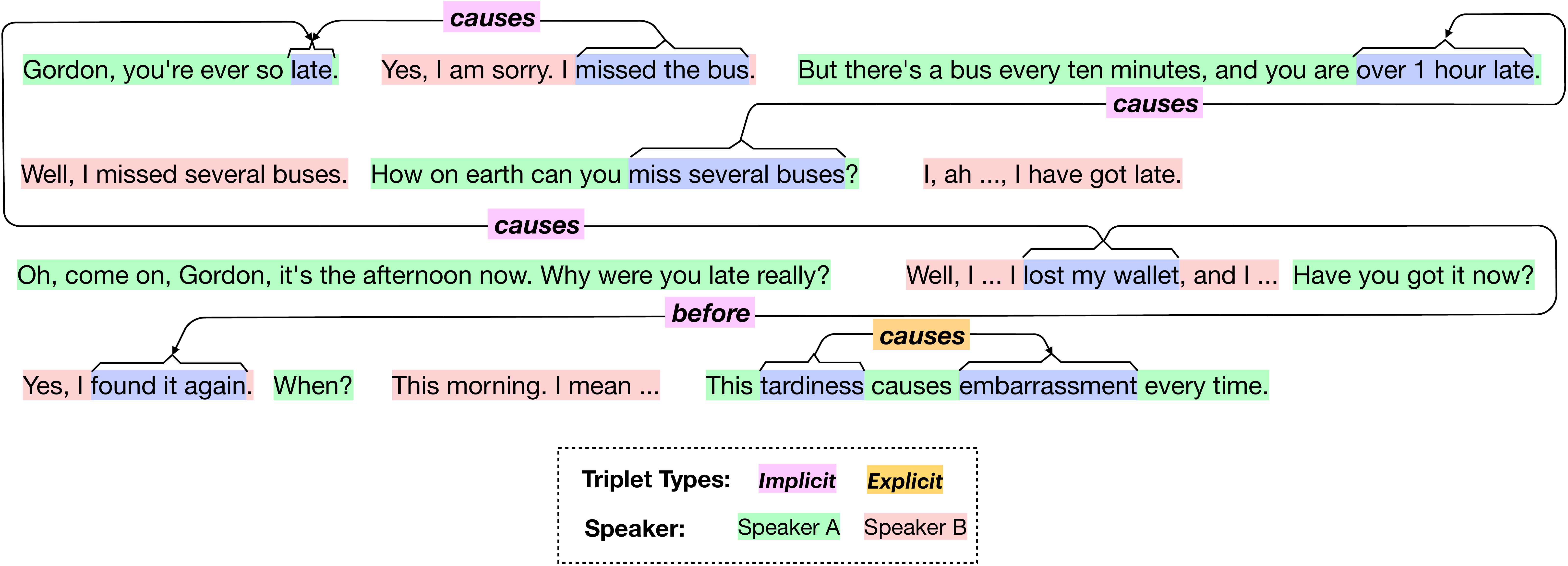}
     \captionof{figure}{ Example of various types of knowledge triplets explaining a dyadic dialogue using commonsense inference; the purple and yellow relations signify implicit and explicit triplets, respectively. }
     \label{fig:teaser}
\end{center}
}]%

\begin{abstract}
Commonsense inference to understand and explain human language is a fundamental research problem in natural language processing. Explaining human conversations poses a great challenge as it requires contextual understanding, planning, inference, and several aspects of reasoning including causal, temporal, and commonsense reasoning. In this work, we introduce {\sc CIDER} -- a manually curated dataset that contains dyadic dialogue explanations in the form of implicit and explicit knowledge triplets inferred using contextual commonsense inference. Extracting such rich explanations from conversations can be conducive to improving several downstream applications. The annotated triplets are categorized 
by
the type of commonsense knowledge present (e.g., causal, conditional, temporal).  We set up three different tasks conditioned on the annotated dataset: Dialogue-level Natural Language Inference, Span Extraction, and Multi-choice Span Selection. Baseline results obtained with transformer-based models reveal that the tasks are difficult, paving the way for promising future research. The dataset and the baseline implementations are publicly available at
\url{https://cider-task.github.io/cider/}.
\end{abstract}

\section{Introduction}
\label{sec:intro}

Understanding and explaining a conversation requires the decomposition of dialogue concepts --- entities, events and actions, and also connecting them through definitive relations. The process of breaking down dialogues into such explanations is grounded in the conversational context and often requires commonsense inference.
Such explanations, when expressed in the form of structured knowledge triplets\footnote{\emph{Knowledge triplets}, and \emph{triplets} are used interchangeably in this paper. In the context of this work, they mean the same.} (\cref{fig:teaser}),
can describe the exact commonsense relation (causal/temporal/conditional/others) through which the concepts are related in the particular conversational context. Establishing such concept links that help explain the dialogue demands two distinct forms of commonsense inference: i) \emph{Explicit} --- the explanation is  verbatim in the triplet. Such triplets %
can be easily extracted out by a parser (e.g., syntactic, pattern matching). These triplets are also prevalent in existing commonsense knowledge graphs~\cite{speer2017conceptnet,sap2019atomic}; and  ii) \emph{Implicit} --- the explanation is entirely contextual, making it more difficult for machines to infer as it requires complex multi-hop commonsense reasoning skills. Our goal is to explain a dialogue by the means of these commonsense inferred triplets. This form of explanation may not be complete, but can give a substantial understanding of the dialogue by breaking it down into contextual triplets. The key element of the dialogue explanation using such triplets is the aspect of contextuality. The triplets extracted from a dialogue using commonsense inference are contextual and are grounded exclusively in that particular dialogue. From our world knowledge, we know that missing a bus \textit{could} cause being late, but (\textit{missed the bus, causes, late}) is grounded and definitive only in the dialogue illustrated in \cref{fig:teaser}. This particular triplet may not be valid in a different dialogue, where the cause of being late could be something different. Similarly, \textit{losing wallet} could cause a different consequence (apart from \textit{being late}) e.g., \textit{getting anxious} in the context of another dialogue. It is also important to highlight that some extracted triplets could be persona-specific. For instance, (\textit{tardiness, causes, embarrassment}) is grounded in the conversation of \cref{fig:teaser}, 
but tardiness may not cause embarrassment for every listener.

In literature, there has been much work on extracting structured knowledge triplets from natural language text. However, there has been only little research to distinguish implicit triplets from explicit triplets present in the text. Explicit triplets can be relatively easily parsed out using semantic parsing~\cite{speer2017conceptnet} and simple co-reference resolution~\cite{joshi2019bert}. Implicit triplets, however, involve non-trivial inference, which becomes even more challenging on dialogue data due to the contextual interplay and latent background knowledge shared between the speakers. Extraction of both explicit and implicit triplets can be conducive to improved dialogue understanding leading to better question-answering systems and richer knowledge bases. To this end, we construct a dataset of 
Commonsense Inference for Dialogue Explanation and Reasoning (CIDER) --
as illustrated in \cref{fig:teaser} -- which captures the relations between  textual concepts or spans appearing in a dialogue. A concept or span can constitute one or multiple entities, objects, actions, states, or events that can be extracted from the dialogue.
The relations are commonsense based, as elaborated in \cref{sec:relations}. Each triplet is tagged as explicit or implicit.

Through this dataset, we aim to evaluate whether state-of-the-art natural language processing models can really read, understand, and comprehend the conversational context of dialogues. We define three tasks on this dataset that require dialogue-level contextual commonsense reasoning --- 
\begin{enumerate*}[label=(\roman*)]
  \item Dialogue-level Natural Language Inference,
  \item Span Extraction, and
  \item Multi-choice Span Selection.
\end{enumerate*}
All three tasks require an overall contextual understanding of the dialogue with commonsense reasoning and inference. 
We setup different state-of-the-art transformer language models as baselines and found that the tasks are challenging to solve.
\noindent\textbf{The Importance of this Dataset:}
The immediate aim of this research is to develop a rich corpus of dialogues with  structured explanations in the form of implicit and explicit triplets, and then use this corpus to perform commonsense inference and reasoning. We formulate non-trivial natural language inference (NLI) and question answering (QA) tasks that can be used to benchmark such reasoning capabilities of natural language processing models.

\begin{figure*}[t]
     \centering
     \begin{subfigure}[b]{0.3\textwidth}
         \centering
         \includegraphics[width=0.92\textwidth]{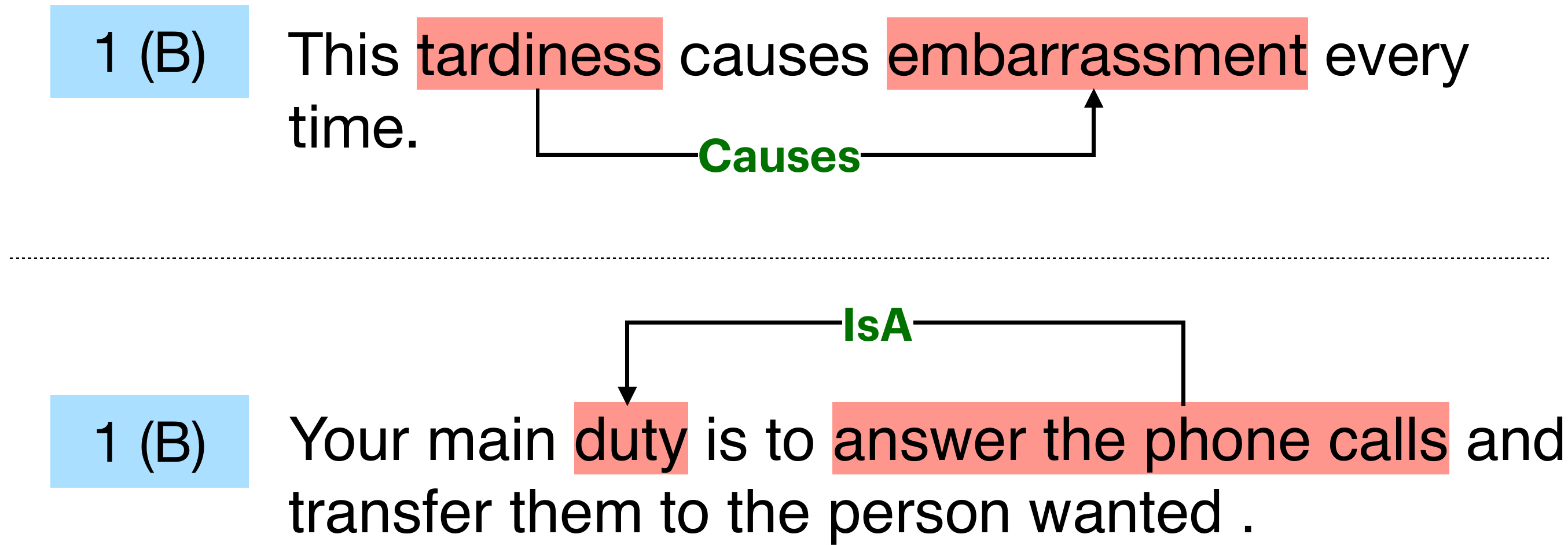}
         \caption{}
         \label{fig:explicit}
     \end{subfigure}
     \hfill
     \begin{subfigure}[b]{0.3\textwidth}
         \centering
         \includegraphics[width=0.92\textwidth]{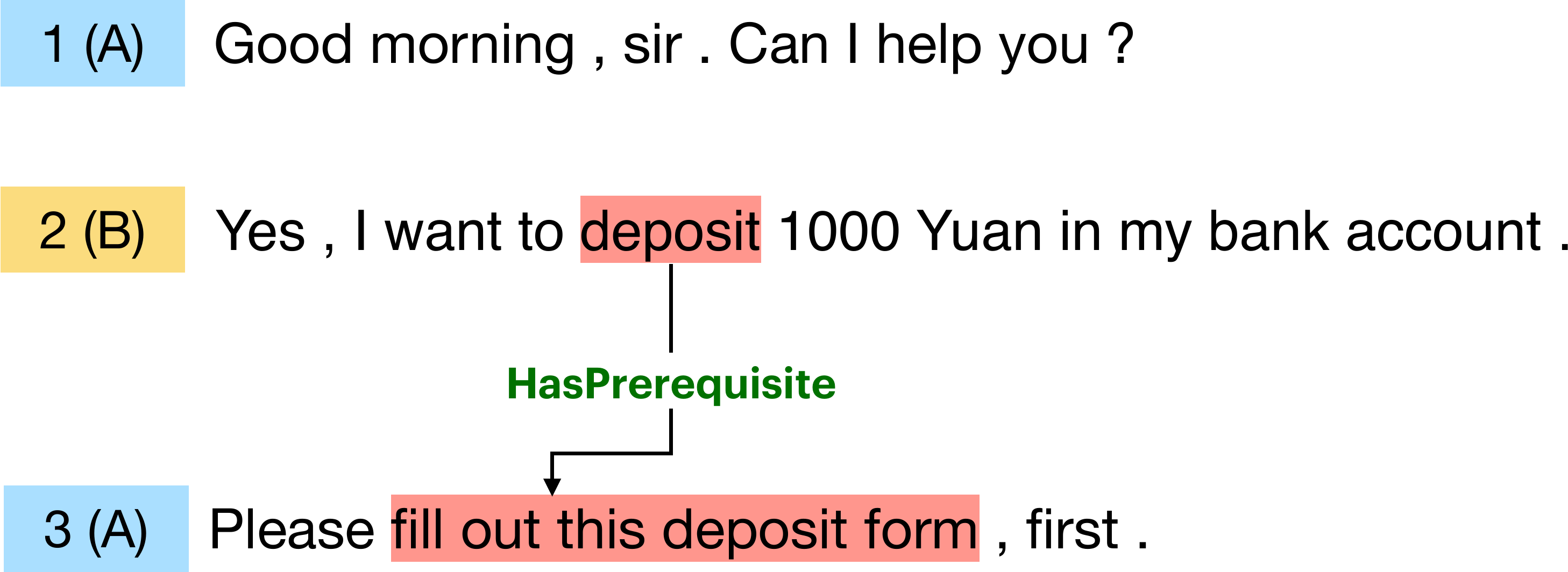}
         \caption{}
         \label{fig:implicit}
     \end{subfigure}
     \hfill
     \begin{subfigure}[b]{0.38\textwidth}
         \centering
         \includegraphics[width=0.91\linewidth]{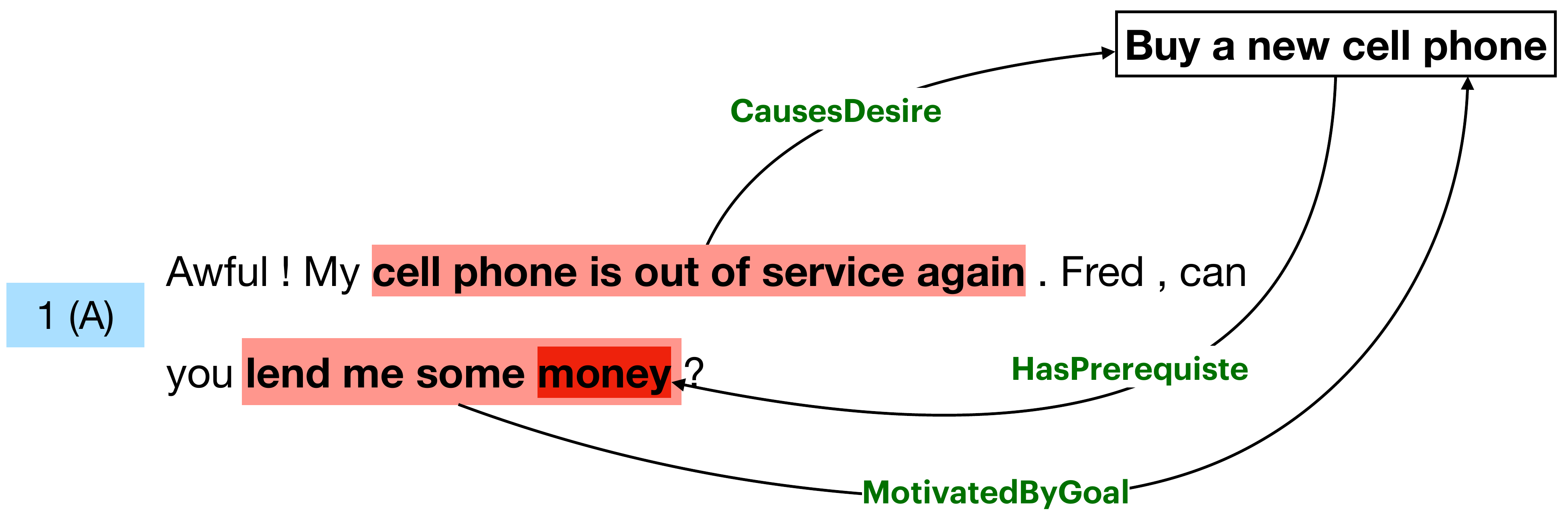}
    \caption{}
     \label{fig:latent}
\end{subfigure}
\label{fig:triplettypes}
      \caption{\footnotesize (a) Explicit and (b) implicit triplets from dialogues. (c) Intermediate latent spans and triplets.}
\end{figure*}

\section{Related Work} 
\label{sec:related_work}
Recently, language models have been scaled up a lot and have seen a performance improvement on various tasks \cite{brown2020language, JMLR:v21:20-074}. However, it has been proved that declarative knowledge is still valuable, especially implicit relationships that are hardly acquired by the state of the art models \cite{hwang2020comet}.

Widely used commonsense knowledge bases such as ConceptNet \cite{speer2017conceptnet} and ATOMIC \cite{sap2019atomic} are
mainly based on crowd-sourced effort.
ConceptNet is a semantic network with nodes composed of common words or phrases in their natural language form. It contains 34 relations, including taxonomic, temporal, and causal ones, such as \textit{MotivatedByGoal} and \textit{Causes}. However, the knowledge in ConceptNet is annotated solely based on the first entity without any other context, making it difficult to capture the long-tail knowledge outside of the most common ones. 
ATOMIC focus on inferential knowledge and consists of nine relations, such as \textit{xIntent} (the intent for personX's action) and \textit{xEffect} (the effect of the event on personX). It covers knowledge around agents involved in the event for if-then reasoning, including subsequent events, mental state, and persona. However, it 
ignores causal relationships between events not carried out by a person. In contrast, our work captures relationships between spans across multiple turns in dialogues. 
As a result of the
dialogue aspect of our data, we also manage to cover implicit knowledge that requires context from conversations to make sense. 

More recent work such as GLUCOSE \cite{mostafazadeh-etal-2020-glucose} which is annotated based on ROCstories \cite{mostafazadeh-etal-2016-corpus} captures implicit knowledge across multiple sentences. %
Our work instead annotates on dialogues, which have more complicated sentences and spoken conversational exchange.

\section{Background}

The primary impetus behind this dataset is
the contextualized structured explanation of a dialogue in the form of 
concept triplets
that can be inferred only through commonsense reasoning. 
The triplets are considered to be the commonsense explanations of different aspects and events that occur in the dialogue. Such aspects would include attributional, comparative, temporal knowledge, and the events may range from physical events involving physical entities, conditional and causal chains, social interactions, persona, etc.

We focus on conversations as our data source, with the choice being motivated by the fact that part of the context in conversations is naturally implicit and interlocutor dependent~\cite{grice75logic}. Commonsense knowledge is considered to be the set of all facts and knowledge about the everyday world which is assumed to be known by all humans \cite{davis2014representations}. For this very reason,  human-to-human dialogues -- typically guided by the Gricean maxims of human interactions -- tend to avoid explicit mentions of commonsense knowledge and the associated reasoning steps.  It is thus reasonable to assume that conversations are generally likely to hold more context-specific inferable implicit knowledge than other genres. This ensures a rich dataset with plenty of contextual implicit triplets
and a reasonable amount of explicit triplets. 

Two distinct spans (e.g., events, entities) in a dialogue may have an implicit connection that can be trivial for humans to interpret using commonsense reasoning and contextual understanding, but can be challenging for machines. %
Uncovering implicit explanations has the potential to enable many important tasks, which we focus on later on. In this work, we propose a dataset that contains manually labeled implicit explanations present in dyadic dialogues that require commonsense reasoning to infer. We use this dataset to evaluate the ability of pre-trained language models' to perform commonsense-based implicit 
reasoning tasks.

The extracted triplets or explanations, of the form $(h, r, t)$ or alternatively $h \xrightarrow{r} t$, consist of a head ($h$) and a tail ($t$) span and the directed relation ($r$) between them. These spans are representative of some events, actions, objects, entities, and so on. The directed relation $r$ comes from a predefined set of relations $\mathcal{R}$ that explain or describe the relationship between the head and tail spans within the context of the conversation --- illustrated in \cref{fig:teaser} with the arrows between spans. Notably, the relation set $\mathcal{R}$ is intended to be generic in nature, rather than specifically factual or taxonomic, so as to accommodate wide categories of knowledge (\cref{sec:relations}) inferred from the context of the conversation. 

\subsection{Types of Triplets}
\label{sec:triplets}

The extracted triplets are either explicit or implicit as defined below:

\textbf{\emph{Explicit triplets}} represent explanations (see \cref{fig:explicit}) that are overtly expressed in an utterance in a dialogue. \cref{fig:teaser} illustrates one such annotated instance in utterance $13$ --- $\textit{tardiness} \xrightarrow{Causes} \textit{embarrassment}$ --- where the triplet is worded verbatim in a head-relation-tail sequence. The head and tail span may contain some pronouns that can be decoded by simple co-reference resolution. In the presence of complex co-reference however the context suggests many possible candidates, and the triplet is implicit.

\textbf{\emph{Implicit triplets}}, on the other hand, are not directly expressed in the dialogue and must be inferable through commonsense reasoning using the contextual information present in the dialogue. Instances of such triplets are shown in \cref{fig:teaser} and \ref{fig:implicit} with the relations in purple font.

\paragraph{Why Focus on Implicit Triplets?}
As pointed out earlier, extracting explicit triplets from a conversation or any natural language text is relatively straightforward and has been studied in much detail in the literature~\cite{auer2007dbpedia,carlson2010toward,speer2017conceptnet}. The much more challenging problem, however, is to extract implicit triplets or explanations.
For example, in \cref{fig:teaser} the triplet $\textit{miss several buses} \xrightarrow{Causes} \textit{over 1 hour late}$ requires commonsense reasoning and knowledge about the world. Similarly, extracting another triplet $\textit{lost my wallet} \xrightarrow{Causes} \textit{late}$ requires multi-utterance reasoning with contextual understanding. Such distillation is not covered by the explicit-triplet extraction framework.
The decomposition of a dialogue into such implicit explanations also requires contextual understanding and complex commonsense reasoning involving multiple steps and utterances. 
Thus, the extraction of implicit explanations is challenging and a focus of this work.

\paragraph{Latent Spans and Differences with GLUCOSE~\cite{mostafazadeh-etal-2020-glucose}:}
As argued earlier, annotating implicit triplets often requires multi-step reasoning. In such cases, one or more intermediate spans (which may not be present in the dialogue) may be required to explain the relation between the constituting spans; see \cref{fig:latent} for one such example. Annotators were given the freedom to identify such intermediate steps when they deemed so. However, such cases are infrequent in our dataset and thus we have chosen to omit the intermediate spans in our experimental studies for the sake of simplicity. 
We leave the intermediate step modelling as a direction for future work.

In this context, it is also important to highlight the fundamental differences between our dataset and GLUCOSE~\cite{mostafazadeh-etal-2020-glucose}. First, in our dataset, the knowledge represented by the spans and the relation connecting them is true (valid) given the context, but establishing this connection using an explicit relation requires complex commonsense inference and understanding of the discourse. The resulting triplet is thus valid in the context and grounded by the context. This is similar to deductive commonsense reasoning \cite{davis2014representations}. GLUCOSE however focuses on abductive commonsense inference, where given an event/state and its context, the annotators provided inferred speculative causal explanations of the event (state) according to {\it their} world and commonsense knowledge. These explanations, although they may fit in the given context, may not always be entailed by it. As a consequence, GLUCOSE is conducive to generative modeling, whereas our dataset leads to extractive modeling.
Second, GLUCOSE has a limited set of relations, where inference is only performed across the following dimensions: \textit{cause}, \textit{enable}, and \textit{result in}. In contrast, we have a much more diverse set of relations
(\cref{sec:relations}). Finally, we construct our dataset based on conversations between two humans, while GLUCOSE is built using monologue-like stories that have significant differences with respect to the discourse structure and semantics.
\subsection{Types of Relations}
\label{sec:relations}
Our proposed \dataset{} dataset contains 25 main and 6 negated relations. Among the main 25 relations, 19 have been adopted from ConceptNet~\cite{speer2017conceptnet}. We introduce 6 new relations to cover some aspects that are not covered by ConceptNet. Brief explanations, examples, and the new relations we introduce are shown in \cref{tab:relations}. 
\begin{table*}[ht!]
  \centering
 \resizebox{0.9\linewidth}{!}{
   \begin{tabular}{p{2cm}|p{3cm}|p{14.1cm}|p{5.3cm}}
   \toprule
   \textbf{Category} & \textbf{Relation} & \textbf{Explanation} & \textbf{Example} \\
   \midrule
   \multirow{7}{*}{Attribution} & Capable Of & Something that A can typically do is B. & knife → cut \\
   & Depends On* & A depends on B. & postage fee → weight of the post \\
   & \multirow{1}{*}{Has A} & \multirow{1}{*}{B belongs to A, either as an inherent part or due to a social construct of possession.} & bird → wing; pen → ink \\
   & Has Property & A has B as a property; A can be described as B. & ice → cold \\
   & Has Subevent & A and B are events, and B happens as a subevent of A. & eating → chewing \\
   & Is A & A is a subtype or a specific instance of B; every A is a B. & car → vehicle; Chicago → city \\
   & Manner Of & A is a specific way to do B. Similar to "Is A", but for verbs. & auction → sale \\
   \midrule
   \multirow{3}{*}{Causal} & Causes & A causes B to happen. & exercise → sweat \\
   & Causes Desire & A makes someone want B. & having no food → buy food \\
   & Implies* & A implies B. & wet cloth → caught in rain \\
   \midrule
   \multirow{7}{*}{Comparison} & \multirow{3}{*}{Antonym} & A and B are opposites in some relevant way, such as being opposite ends of a scale, or fundamentally similar things with a key difference between them. Counter-intuitively, two concepts must be quite similar before people consider them antonyms.  & \multirow{3}{*}{black $\xleftrightarrow{}$ white; hot $\xleftrightarrow{}$ cold} \\
   & \multirow{1}{*}{Distinct From} & A and B are distinct member of a set; something that is A is not B. & \multirow{1}{*}{red $\xleftrightarrow{}$ blue; August $\xleftrightarrow{}$ September} \\
   & Similar To & A is similar to B. & mixer $\xleftrightarrow{}$ food processor \\
   & \multirow{2}{*}{Synonym} & A and B have very similar meanings. They may be translations of each other in different languages. & \multirow{2}{*}{sunlight $\xleftrightarrow{}$ sunshine} \\
   \midrule
   \multirow{1}{*}{Conditional} & \multirow{1}{*}{Has Prerequisite} & In order for A to happen, B needs to happen; B is a dependency of A. & \multirow{1}{*}{dream → sleep} \\
   \midrule
   \multirow{5}{*}{Intentional} & \multirow{2}{*}{Desires} & A is a conscious entity that typically wants B. Many assertions of this type use the appropriate language's word for "person" as A. & \multirow{2}{*}{person → love} \\
   & \multirow{1}{*}{Motivated By Goal} & Someone does A because they want result B; A is a step toward accomplishing the goal B. & \multirow{1}{*}{compete → win} \\
   & \multirow{1}{*}{Obstructed By} & A is a goal that can be prevented by B; B is an obstacle in the way of A. & \multirow{1}{*}{sleep → noise} \\
   & Used For & A is used for B; the purpose of A is B. & bridge → cross water \\
   \midrule
   \multirow{1}{*}{Social} & Social Rule* & A is the social norm for when B happens or during B. & apology → late  \\
   \midrule
   \multirow{2}{*}{Spatial} & At Location & A happens at location B, or B is a typical location for A. & try clothes → changing room  \\
   & Located Near & A and B are typically found near each other. & table → chairs  \\
   \midrule
   \multirow{3}{*}{Temporal} & Before* & A starts/ends before B. & brush teeth → go to bed \\
   & Happens On* & A happens during B. & celebration → birthday \\
   & Simultaneous* & A and B happens at the same time. & heavy sports → heavy breath \\
   \bottomrule
   \end{tabular}
   }
  \caption{\footnotesize Annotated relations in our dataset. * indicates new relations introduced by us that are not present in ConceptNet. $\xleftrightarrow{}$ in the examples indicate symmetric relations. In addition to the above, we also have a few negation relations as illustrated in \cref{sec:negrel}. }
  \label{tab:relations}
\end{table*}
We categorize the different relations as follows:

\noindent\textbf{Attribution.}
Relations that indicate attributes, properties, and definitions of concepts:
\begin{enumerate*}[label=(\roman*)]
  \item \textit{Capable Of},
  \item \textit{Depends On},
  \item \textit{Has A},
  \item \textit{Has Property},
  \item \textit{Has Subevent},
  \item \textit{Is A}, and
  \item \textit{Manner Of}
\end{enumerate*}.

\noindent\textbf{Causal.} Relations the indicate cause and effect of events: \begin{enumerate*}[label=(\roman*)]
  \item \textit{Causes},
  \item \textit{Causes Desire}, and
  \item \textit{Implies}
\end{enumerate*}.

\noindent\textbf{Comparison.} Relations that indicate comparison, similarity, or dissimilarity between concepts: \begin{enumerate*}[label=(\roman*)]
  \item \textit{Antonym},
  \item \textit{Distinct From},
  \item \textit{Similar To}, and
  \item \textit{Synonym}.
\end{enumerate*}

\noindent\textbf{Conditional.}
This category, having only one relation \textit{Has Prerequisite},
indicates dependency of one event on the other.

\noindent\textbf{Intentional.}
Relations indicating intent or usage of an entity or a person: \begin{enumerate*}[label=(\roman*)]
  \item \textit{Desires},
  \item \textit{Motivated By Goal},
  \item \textit{Obstructed By}, and
  \item \textit{Used For}.
\end{enumerate*}

\noindent\textbf{Social.}
The category involves social commonsense relations specifying social rules, conventions, norms, and suggestions. The relation in this category is:
\begin{enumerate*}[label=(\roman*)]
  \item \textit{Social Rule}.
\end{enumerate*}

\noindent\textbf{Spatial.}
This category encompasses relations which signifies spatial properties, such as location of events, entities, actions. The relations include:
\begin{enumerate*}[label=(\roman*)]
  \item \textit{At Location}, and
  \item \textit{Located Near}.
\end{enumerate*}

\noindent\textbf{Temporal.} This category involves the idea of time considering the start, end, duration, and order of events. The constituent relations are:
\begin{enumerate*}[label=(\roman*)]
  \item \textit{Before},
  \item \textit{Happens On}, and
  \item \textit{Simultaneous.}
\end{enumerate*}

\subsection{Negative and Symmetric Relations}
\label{sec:negrel}
Apart from the relations in \cref{tab:relations}, the negations of some of  these relations are necessary to form the triplets during annotation. These negated relations are
\begin{enumerate*}[label=(\roman*)]
    \item \textit{Not Causes},
    \item \textit{Not Causes Desire},
    \item \textit{Not Has Property},
    \item \textit{Not Implies}, 
    \item \textit{Not Is A}, and
    \item \textit{Not Motivated By Goal}.
\end{enumerate*}

It should be noted that there are some symmetric relations in our relation set. A relation $R$ is considered symmetric if the validity of $A \xrightarrow{R} B$ implies the validity of $B \xrightarrow{R} A$ and vice versa. The set of symmetric relations $\bm{\mathcal{R^S}}$ contains
\begin{enumerate*}[label=(\roman*)]
  \item \textit{Antonym},
  \item \textit{Distinct From},
  \item \textit{Similar To},
  \item \textit{Synonym},
  \item \textit{Located Near}, and
  \item \textit{Simultaneous}.
\end{enumerate*}

A few more negative relations were annotated in our dataset, but was not considered in our experiments due to their very less frequency.

\section{Dataset Construction}
\label{sec:dataset}

\subsection{Source Datasets of Dialogues}
The annotation is performed on the following datasets containing dyadic dialogues:

\noindent \textbf{DailyDialog}~\cite{li2017dailydialog} is aimed towards emotion and dialogue-act classification at utterance level. The conversations cover various topics ranging from ordinary life, work, and relationships, to tourism, finance and politics. 

\noindent \textbf{MuTual}~\cite{mutual} is a manually annotated dataset for multi-turn dialogue reasoning. It was introduced to evaluate several aspects of dialogue-level reasoning in terms of next utterance prediction given a dialogue history. These aspects include attitude reasoning, intent prediction, situation reasoning, multi-fact reasoning, and others.

\noindent  \textbf{DREAM}~\cite{sun2019dream} is a dialogue-based multiple-choice reading-comprehension dataset collected from exams of English as a foreign language. This dataset presents several challenges as it contains non-extractive answers that require commonsense reasoning beyond a single sentence.

In total, we sampled 807 dialogues from the three datasets. Each sampled dialogue has 5 to 12 utterances,  and each constituent utterance has no more than 30 words. 

\subsection{Annotation Process}
\label{sec:annot}

\paragraph{Annotation guidelines.}
The annotators are instructed to identify either explicit or implicit triplets in a dialogue 
(\cref{sec:triplets}). Such a triplet consists of a pair of spans $A$ and $B$, and an appropriate relation $R$ between them, denoted as $A \xrightarrow{R} B$. A \emph{span} is defined as a word, phrase, or a sub-sentence unit of an utterance that represents an entity, event, concept or action.
The annotators are instructed to meet the following constraints during the annotation:
\begin{itemize}[leftmargin=*, wide, itemsep=0em, labelwidth=!, labelindent=0pt]
    \item The extracted triplets must be entailed by the conversation to be valid.

    \item The spans of a triplet should be as short and concise as possible. 
    Also, a triplet may connect a pair of spans from distinct utterances in a dialogue.
    
    \item Multiple distinct valid relations between the same pair of spans are allowed. All these relations correspond to distinct triplets.
\end{itemize}

We used a web-based tool called BRAT~\cite{brat} for the annotation. The annotators are three PhD students who have thorough knowledge about the task. They were first briefed about the annotation rules, followed by a trial with a few samples to evaluate their understanding of the annotation guidelines and ability to extract both explicit and implicit triplets. Although annotators extract both types, they were instructed to focus more on annotating implicit triplets since extracting those are more challenging. The trial stage was conducted to ensure that annotators are well-versed in annotating high quality triplets in the final phase.

\subsection{Annotation Verification and Agreement}

Each dialogue is primarily annotated by a single annotator. We then verify the validity of the annotated triplets using the following strategy:
\begin{enumerate}[leftmargin=*, wide, itemsep=0em, labelwidth=!, labelindent=0pt]
    \item All extracted triplets are independently validated by two other validation annotators, in terms of their inferability from their source dialogues.
    \item Unanimously agreed-upon valid  triplets are kept, while unanimously agreed-upon invalid triplets are discarded. In the case of a disagreement, we bring in a third annotator to break the tie. 
    \item The final set of valid triplets is labelled as being explicit or implicit by the same two annotators as in step (1). The majority vote is assigned as the final label. Similar to the previous step, in case of a disagreement, we bring in a third annotator to break the tie. 
\end{enumerate}
\noindent After this stage, we obtained a Cohen's Kappa inter-validation-annotator agreement of 0.91 for triplet verification and 0.93 for 
relation type labelling. 
We found that the number of explicit triplets (4.5\%) in the final annotated dataset is significantly less than implicit triplets (95.5\%). The reason is the informal nature of the source datasets' conversations, which enables the extraction of much more frequent implicit triplets than explicit ones. Statistics of the annotated dataset are shown in \cref{tab:stat}.

\begin{table}[ht!]
\small
\centering
\resizebox{0.9\linewidth}{!}{
	\begin{tabular}{p{8.4cm}@{}|r@{~~}}
	\toprule
	\textbf{Description} & \textbf{Instances} \\
	\midrule
    \# Dialogues/\# triplets in DailyDialog & 245/1286 \\
    \# Dialogues/\# triplets in MuTual & 182/658 \\
    \# Dialogues/\# triplets in DREAM & 380/2595 \\
    \# Dialogues/\# triplets Total & 807/4539 \\
    \midrule
    \# Dialogues with \# triplets $<$ 3 & 142 \\
    \# Dialogues with \# triplets between 3-5 & 312 \\
    \# Dialogues with \# triplets between 5-10 & 281 \\
    \# Dialogues with \# triplets $>$ 10 & 72 \\
    Average \# triplets per dialogue & 5.62 \\
    \midrule
    \# Triplets with spans from Utt. distance = 0 & 1009 \\
    \# Triplets with spans from Utt. distance = 1 & 1490 \\
    \# Triplets with spans from Utt. distance between 2-5 & 1501\\
    \# Triplets with spans from Utt. distance between 6-8 & 401\\
    \# Triplets with spans from Utt. distance $>$ 8 & 138\\
    \midrule
    \# Triplets having spans from same speaker & 2475 \\
    \# Triplets having spans from different speakers & 2064 \\
    \midrule 
    \# Span pairs with single relation & 4203 \\
    \# Span pairs with multiple relations & 164 \\
    \bottomrule
	\end{tabular}
	}
	\caption{\footnotesize{Statistics on our dataset \dataset{}. Please refer to the appendix for frequency statistics of the relations.}}
	\label{tab:stat}
\end{table}

\section{Experimental Setup and Results} 
\label{sec:experiments}

We formulate three  tasks on the \dataset{} dataset: 1) Dialogue-level Natural Language Inference; 2) Span Extraction;  3) Multi-choice Span Selection.

\subsection{Dialogue-level Cross Validation}
\label{sec:dataprep}
We consider a dialogue-level cross-validation strategy to benchmark our models. We partition the annotated dialogues into five disjoint and roughly equal-sized folds. Per cross-validation round, the triplets from four folds are considered for training, and the remaining one fold is used for test. 

\subsection{Task 1: Dialogue-level Natural Language Inference (DNLI)}
\label{sec:dnli}

\begin{figure*}[t]
     \centering
     \begin{subfigure}[t]{0.49\linewidth}
         \centering
         \includegraphics[width=0.7\textwidth]{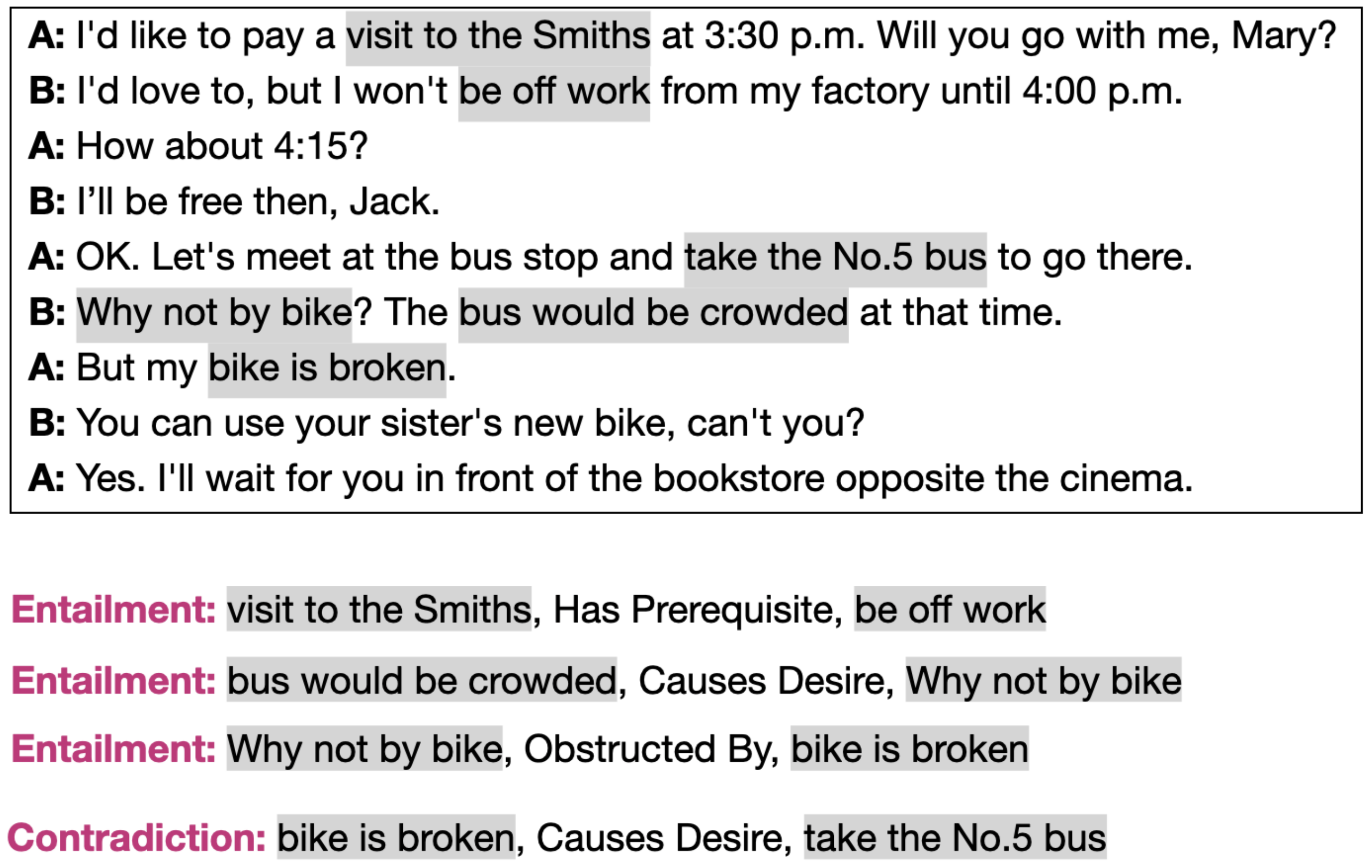}
         \caption{}
         \label{fig:subtask1}
     \end{subfigure}
     \hfill
     \begin{subfigure}[t]{0.49\linewidth}
         \centering
         \includegraphics[width=0.7\textwidth]{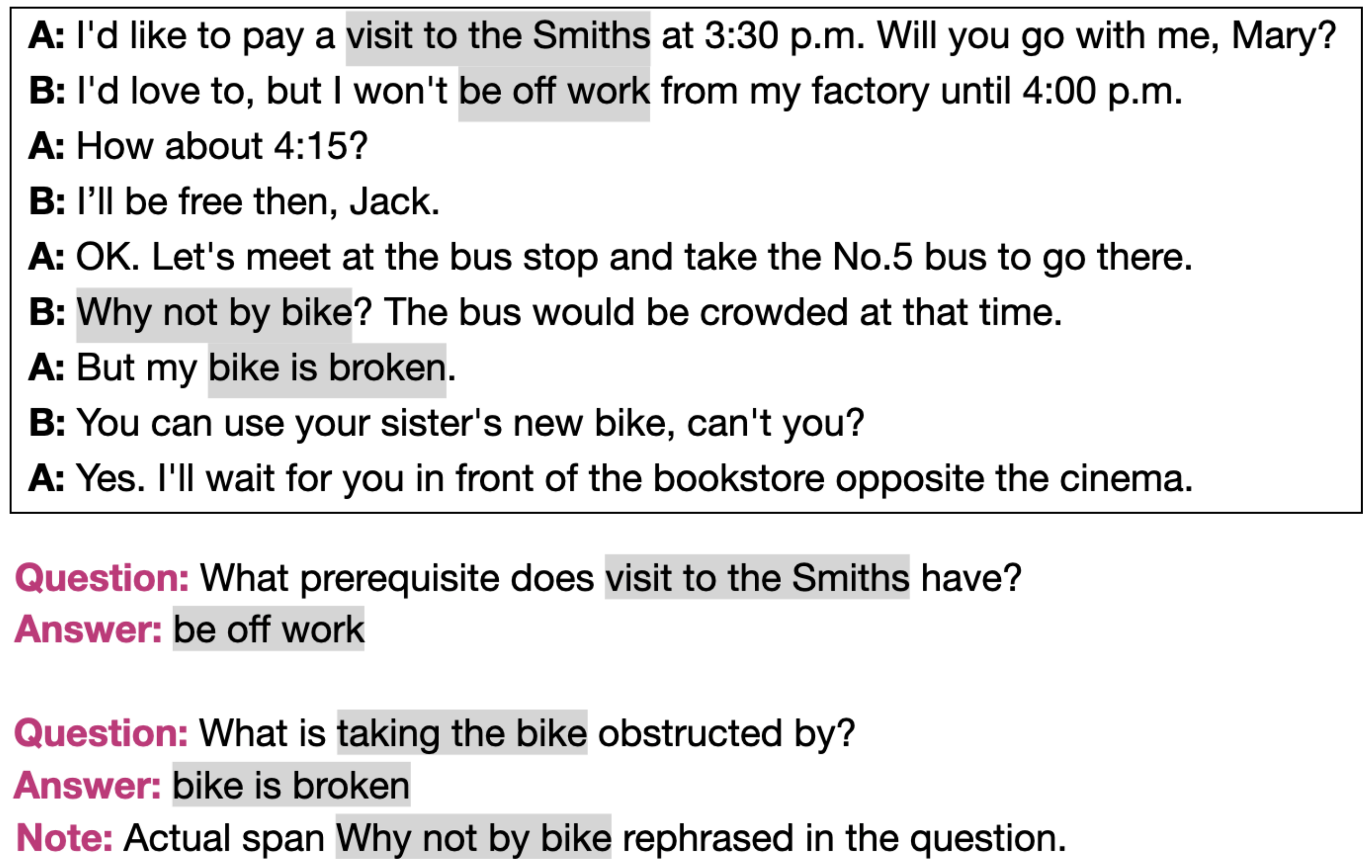}
         \caption{}
         \label{fig:subtask2}
     \end{subfigure}
    \hfill
   \caption{\footnotesize (a) Subtask 1: Dialogue-level Natural Language Inference (DNLI). (b) Subtask 2: Span Extraction.}
\end{figure*}

Textual entailment, later renamed as natural language inference (NLI), is the task of identifying if a ``hypothesis'' is true (entailment), false (contradiction), or undetermined (independent) given a ``premise''. We extend this definition to conversations and propose \textit{Dialogue-level Natural Language Inference} (DNLI), which is the task of determining whether a triplet (hypothesis) is true or false given a dialogue (premise) (see \cref{fig:subtask1}). 

It should be noted that most NLI datasets such as SNLI~\cite{bowman2015large}, MultiNLI~\cite{williams2017broad}, SciTail~\cite{khot2018scitail} consist of a single sentence hypothesis and premise, whereas for DNLI the hypothesis and the premise are a triplet and a conversation, respectively. 

For our experiments, the \emph{hypothesis} is formed by concatenating and lemmatizing the elements of the triplet $h \xrightarrow{r} t$ in $h,r,t$ order. Lemmatization is performed to remove surface level grammatical clues from the triplet. The \emph{premise} is formed by concatenating the utterances of the dialogue.

\subsubsection{Creating Negative Examples}
\label{sec:neg}
Let $C$ be a conversation, $T$ be the set of all valid triplets in $C$, and $A \xrightarrow{R} B$ be one such valid triplet in $T$. We denote $\bm{\mathcal{R}}$: set of all relations; $\bm{\mathcal{R^S}}$: set of symmetric relations. The samples with valid triplets as hypotheses are termed as positive examples. The contradicting triplets/hypotheses for the negative samples are created from $T$ as follows:

\begin{table}[ht!]
\small
\centering
\resizebox{0.9\linewidth}{!}{
\begin{tabular}{ll|ccccc}
\toprule
\textbf{Split} & \textbf{Label} & \textbf{Fold1} & \textbf{Fold2} & \textbf{Fold3} & \textbf{Fold4} & \textbf{Fold5} \\
\toprule
\multirow{2}{*}{Train} & Positive & 3504 & 3510 & 3508 & 3507 & 3515 \\
& Negative & 6195 & 6223 & 6281 & 6246 & 6224 \\
\midrule
\multirow{2}{*}{Test} & Positive & 882 & 876 & 878 & 879 & 871 \\
& Negative & 7185 & 6636 & 6749 & 6821 & 6947 \\
\bottomrule
\end{tabular}
}
\caption{\footnotesize Cross validation fold statistics for Task 1: DNLI.}
\label{tab:nli_stat}
\end{table}

\noindent\textbf{Reverse Relation Direction.}
In $A \xrightarrow{R} B$, if $R \notin \bm{\mathcal{R^S}}$, then $B \xrightarrow{R} A$ is a contradicting hypothesis.

\noindent\textbf{Substitute Relation Type.}
For $A \xrightarrow{R} B$, another relation $Q$ is randomly sampled from $\bm{\mathcal{R}} \setminus \{R\}$ and $A \xrightarrow{Q} B$ is considered a contradicting hypothesis. 

\noindent\textbf{Substitute Span.}
For $A \xrightarrow{R} B$, either $A$ or $B$ is replaced with another random span $X$ from the other triplets in set $T$. $X \xrightarrow{R} B$ or $A \xrightarrow{R} X$ is then considered a contradicting hypothesis.

\noindent\textbf{Combination of All.} A combination of the above three strategies can also be used to create the contradicting hypothesis. We ensure that the contrived contradicting hypotheses do not appear in the set of annotated triplets $T$.

The above strategies allow us to create multiple negative samples from a positive sample. In our experiments, we had two and eight negative samples per positive sample in the training and test split, respectively. We intentionally keep fewer negative samples in the training data to evaluate the generalization capacity of the models on a more diverse range of negative samples in the test data. Fold-wise statistics are shown in \cref{tab:nli_stat}. An example of the DNLI task is illustrated \cref{fig:subtask1}.

\subsubsection{Baseline}

\paragraph{RoBERTa-large Fine-tuned on MNLI.} We use the pretrained \code{roberta-large-mnli} model~\cite{liu2019roberta} to benchmark this task. The input to the model is: \code{<CLS> Premise <SEP> Hypothesis <SEP>}. The classification is performed on the \code{<CLS>} token vector from the final layer. We choose this model as it has been fine-tuned on the MNLI dataset and shows impressive performance on a number of NLI tasks.

The performance of the RoBERTa-MNLI model is reported in \cref{tab:nli}. As DNLI is a classification task, we report macro F1, weighted F1, and precision and recall over the positive examples (with valid triplets). We notice that the metrics are quite consistent across the five different folds and thus we report our conclusion against the average score. 
We obtained an average weighted F1 score of \textbf{85.78\%}. However, the macro F1 score is noticeably lower at \textbf{69.83\%}, suggesting that the model performs poorly on the less-frequent positive examples. 
The recall score suggests that \textbf{76.85\%} of the valid hypotheses are correctly identified by the model. However, the precision score is quite low at \textbf{37.25\%}, suggesting that almost $2/3$-rd of the predicted valid hypothesis are in-fact invalid. Without fine-tuning, the model produces much lower macro F1 of \textbf{17.76\%}, precision of \textbf{15.06\%}, and recall of \textbf{47.4\%}. The state-of-the-art RoBERTa MNLI model is thus not very capable of correctly identifying triplets entailed by the conversation. We conclude that 
inference from conversational context based on commonsense reasoning is not straightforward for pretrained language models.

\subsection{Task 2: Span Extraction}
\label{sec:se}
\textit{Span Extraction} is defined as identifying the tail span $B$, given the head span $A$, the relation $R$ between $A$ and $B$, and the conversation $C$ where $A \xrightarrow{R} B$ is encoded.
It is analogous to the task of node prediction in knowledge bases, where the missing tail node $B$ in $A \xrightarrow{R} ?$ is to be predicted. 
\cref{fig:subtask2} depicts an example of this subtask.

\begin{table}[t]
\small
\centering
\resizebox{0.9\linewidth}{!}{

\begin{tabular}{l|ccccc|c}
\toprule
\textbf{Metric} & \textbf{Fold1} & \textbf{Fold2} & \textbf{Fold3} & \textbf{Fold4} & \textbf{Fold5} & \textbf{Avg.} \\
\midrule
Macro F1 & 69.15 & 71.07 & 68.14 & 71.29 & 69.49 & 69.83 \\
Weighted F1 & 86.76 & 85.48 & 84.07 & 86.42 & 86.17 & 85.78 \\
Precision Positive & 35.79 & 39.18 & 34.87 & 39.37 & 37.05 & 37.25 \\
Recall Positive & 77.55 & 78.54 & 77.56 & 78.16 & 72.45 & 76.85 \\

\bottomrule
\end{tabular}
}
\caption{\footnotesize Results for the RoBERTa-MNLI model in Task 1: Dialogue-level Natural Language Inference (DNLI).}
\label{tab:nli}
\end{table}

\multirow{5}{*}{\rotatebox{90}{\small{}}}

In this paper, \textit{Span Extraction} is formulated as a Machine Reading Comprehension (MRC) task similar to
SQuAD~\citep{rajpurkar2016squad} where a question is to be answered from a given passage of text or more generally context. The equivalencies with MRC are defined as follows:

\noindent \textbf{Context.} The entire conversation $C$ is treated as the context, as the span $B$ in the triplet $A \xrightarrow{R} B$ can come from any utterance of $C$.

\noindent \textbf{Question and Answer.} For each relation type $R$, we create a question template that includes a placeholder for span $A$ and asks for span $B$ as the answer. The templates are filled with the appropriate valid triplets to generate the question-answer pairs.
Please refer to the question template in appendix.

\subsubsection{Baselines}
We use two pretrained transformer-based models to benchmark the \textit{Span Extraction} task. The methodology described in BERT QA models~\cite{devlin2018bert} is used to extract the tail-spans/answers.\\
\noindent\textbf{RoBERTa Base.} We use the \code{roberta-base} model~\cite{liu2019roberta} as a baseline model. 
\noindent\textbf{SpanBERT Fine-tuned on SQuAD.} We use SpanBERT~\citep{joshi2020spanbert} fine-tuned on SQuAD 2.0 dataset as the other baseline model. 
\subsubsection{Evaluation Metrics}
\noindent \textbf{EM (Exact Match).} \% of the predicted answers that are identical to the gold answers. 
\textbf{NM (No Match).} \% of the predicted answers that bear no match with the gold answer. \textbf{F1:} The F1 score introduced by \citet{rajpurkar2016squad} to evaluate word-level overlap of predictions with the gold answers for extractive QA models.

\subsubsection{Results}

\noindent The results for this task is reported in \cref{tab:span}. We notice that the SpanBERT model performs significantly better than the RoBERTa model. This is expected as SpanBERT has been pretrained with a different objective function and it particularly excels at span extraction tasks, such as, question answering. However, the EM score of \textbf{28.41\%} and the F1 score of \textbf{42.06\%} for the superior SpanBERT model is still subpar. The EM score suggests that the model extracts the exact correct answer less than $1/3$-rd of the time. The NM score also indicates that the extracted answer and the actual answer have no overlap around half of the time. Without fine-tuning, the SpanBERT model produces an EM score of \textbf{7.96\%} and a F1 score of \textbf{20.78\%}, much lesser than the fine-tuned model. We conclude that the state-of-the-art pretrained language models struggle with extracting missing spans.
\begin{table}[t]
\small
\centering
\resizebox{0.9\linewidth}{!}{
\begin{tabular}{ll|ccccc|c}
\toprule
\textbf{Model} & \textbf{Metric} & \textbf{Fold1} & \textbf{Fold2} & \textbf{Fold3} & \textbf{Fold4} & \textbf{Fold5} & \textbf{Avg.} \\
\midrule

\multirow{3}{*}{SpanBERT} & EM & 29.2 & 28.35 & 26.57 & 31.54 & 26.37 & 28.41 \\
& NM & 46.47 & 48.71 & 52.91 & 47.48 & 50.0 & 49.11 \\
& F1 & 43.72 & 42.27 & 39.31 & 44.22 & 40.77 & 42.06 \\
\midrule
\multirow{3}{*}{RoBERTa} & EM & 15.87 & 13.18 & 12.1 & 15.12 & 13.48 & 13.95 \\
& NM & 57.36 & 56.71 & 61.57 & 53.22 & 57.4 & 57.25 \\
& F1 & 31.31 & 30.83 & 28.93 & 34.38 & 31.86 & 31.46 \\

\bottomrule
\end{tabular}
}
\caption{\footnotesize Results for Span Extraction task. Higher EM, F1, and lower NM scores are better. 
}
\label{tab:span}
\end{table}
\subsection{Task 3: Multi-choice Span Selection}

\label{sec:mss}
\textit{Multi-choice Span Selection} is motivated by the SWAG commonsense inference task~\cite{zellers2018swag}. In SWAG, given a partial description of a situation, the appropriate ending is to be selected from a given list of choices using commonsense inference. In our case, \textit{Multi-choice Span Selection} is formulated as a multiple-choice question answering task. Similar to the previous task, given a conversation $C$ and partial information about a triplet $A \xrightarrow{R} ?$, the goal is to predict the missing span $B$ as an answer to a question created from $A$ and $R$. However, in contrast to task 2, the missing span $B$ has to be selected from a list of four possible answers $S = \{s_1, ..., s_4\}$. We show an example of this task in \cref{fig:subtask3}.
The context, question, and answers for this task are created as follows:

\begin{figure}[ht!]
         \centering
         \includegraphics[width=0.8\linewidth]{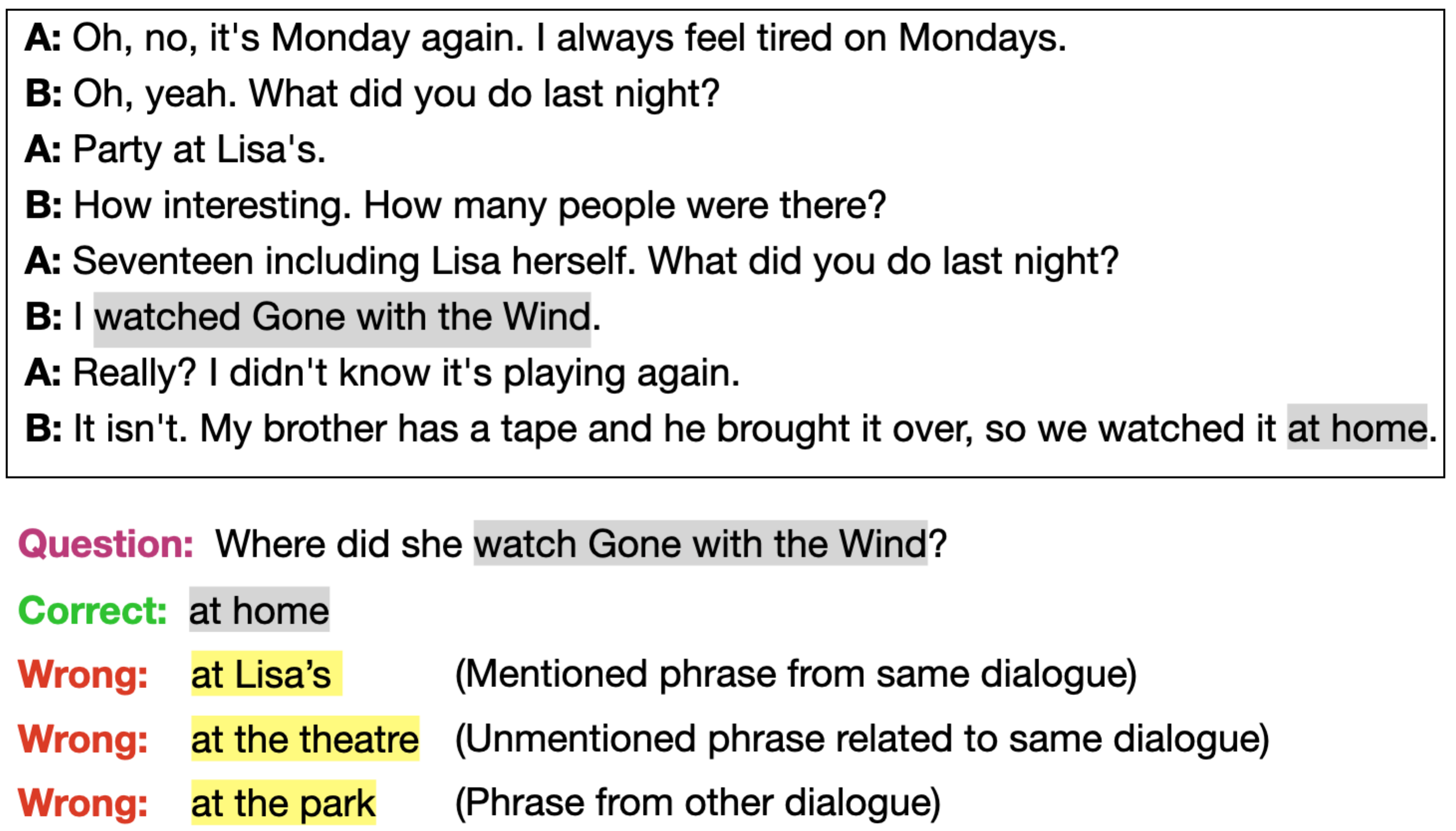}
     \caption{\footnotesize Subtask 3: Multi-choice Span Selection.}
     \label{fig:subtask3}
\end{figure}

\noindent \textbf{Context and Question:} Both the context and the question construction follow \cref{sec:se}. 

\noindent \textbf{Correct and Confounding Options:} 
The options include the target answer 
and the three confounding options that are extracted from the same context 
.

\subsubsection{Creating Confounding Options}
\label{sssec:creating_confounding_options}
To mitigate the stylistic artifacts that could give away the target answer~\citep{gururangan2018annotation, poliak2018hypothesis}, the confounding options are generated in an adversarial fashion.

\paragraph{Generating Confounding-option Candidates.} 
We first select a large number of spans from $C$ to form a confounding-option collection $\mathcal{N}$ by leveraging the SpanBERT
fine-tuned on the samples of Task 2~(\cref{sec:se}). We feed each individual utterance as the context, and the question created from $A$ and $R$ to the SpanBERT fine-tuned for Task-2. This leads to one or two candidate answers (spans) per contextual utterance per question, averaging around 30 confounding spans per question. We discard the spans that form a valid triplet with $A$ and $R$.

\paragraph{Adversarial Filtering.} Once we have the collection $\mathcal{N}$, we follow \citet{zellers2018swag}  to filter the
confounding options generated in \cref{sssec:creating_confounding_options}.
Please check Appendix Section A for more details. We use the
\code{roberta-base} model to filter out stylistic patterns. During the filtering process, discriminator
prediction accuracy decreased from 0.55 to 0.27,
suggesting the method’s effectiveness in removing
easy confounding candidates with stylistic patterns.

\subsubsection{Baseline}
We experiment with \code{bert-base-uncased} and \code{roberta-base} on the adversarially created dataset. The input to the models is the concatenation of conversation $C$, question $Q$, and candidate answers $A_j, j \in \{1,...,4\}$: \code{<CLS> C <SEP> Q <SEP> A\_j <SEP>}. Each score is predicted from the corresponding \code{<CLS>} token vector and the highest scoring one is selected as answer.

\subsubsection{Results}
The results reported in \cref{tab:mss1} indicate the importance of contextual information in improving models' performance. Our human verifiers could also predict the answers significantly more accurately when contextual information was available. 
It is worth noting that all the pre-trained language models perform poorly in this task and the obtained results are far from reaching the human-level performance. 
Besides, the accuracy score for \code{bert-base-uncased} and \code{roberta-base} without fine-tuning  are 25.60\% and 26.22\% respectively which is similar to a random baseline (25.00\%), confirming the conclusion in Task 2~(\cref{sec:se}) that current language models have difficulties in predicting the missing span.
\begin{table}[t]
\centering
\small
\resizebox{0.9\linewidth}{!}{
\begin{tabular}{ll|ccccc|c}
\toprule
\textbf{Model} & \textbf{Setting} & \textbf{Fold1} & \textbf{Fold2} & \textbf{Fold3} & \textbf{Fold4} & \textbf{Fold5} & \textbf{Avg.} \\ \midrule
\multirow{2}{*}{BERT} & C\&Q & 60.35 & 58.96 & 51.84 & 61.62 & 60.55 & 58.66 \\
 & Q & 47.21 & 50.89 & 51.25 & 54.46 & 47.84 & 50.33 \\
 \midrule
\multirow{2}{*}{RoBERTa} & C\&Q & 61.16 & 51.05 & 65.28 & 73.31 & 62.04 & 62.57 \\
 & Q & 51.05 & 62.04 & 56.60 & 58.92 & 55.76 & 56.87 \\
 \midrule
\multirow{2}{*}{Human} & C\&Q & 89.90 & 82.69 & 83.02 & 80.77 & 80.78 & 83.43 \\
 & Q & 69.39 & 67.31 & 60.00 & 65.38 & 71.15 & 66.45 \\ 
 \bottomrule
\end{tabular}
}
\caption{\footnotesize Results for Multi-choice Span Selection. C\&Q $\xrightarrow{}$ model input is Context, Question; Q $\xrightarrow{}$ input is only Question.}
\label{tab:mss1}
\end{table}
\begin{table}[ht!]
\centering
\small
\resizebox{0.8\linewidth}{!}{

\begin{tabular}{lccc}
\toprule
\textbf{Relation Type} & \textbf{Subtask 1} & \textbf{Subtask 2} & \textbf{Subtask 3} \\
\midrule
Attribution & 74.97 & 43.34 & 64.64 \\
Causal & 67.26 & 38.04 & 61.20 \\
Comparison & 68.75 & 36.78 & 58.76 \\
Conditional & 68.51 & 38.97 & 55.72 \\
Intentional & 70.49 & 46.70 & 63.34 \\
Social & 58.97 & 28.34 & 58.00 \\
Spatial &79.06 & 57.41 & 71.20 \\
Temporal & 71.56 & 54.26 & 54.53 \\
\bottomrule
\end{tabular}
}
\caption{\footnotesize Average five-fold Macro-F1, F1, and Accuracy score over the relation categories.
We report results for RoBERTa-MNLI, SpanBERT and RoBERTa models for the three tasks.
}
\label{tab:relation-all}

\end{table}

\paragraph{Performance across Relation Categories.}
We report the results across different relation categories for each task with the corresponding best performing models in \cref{tab:relation-all}. We notice that \textit{Spatial} is one of the top-performing categories across all three tasks. Performance in \textit{Attribution} and \textit{Temporal} category are also reasonably well in Task 1 and Task 1, 2 respectively. Interestingly, the result of \textit{Temporal} category in Task 3 is the worst. The performance in \textit{Causal} and \textit{Conditional} category is around the average mark across all three tasks. This implies that pretrained language models find it difficult to understand the concept of causal events or dependent events. Finally, we observe that the performance in \textit{Social} category is the worst or among the worst for all the tasks, suggesting that the models find it very challenging to reason about social norms, rules, and conventions.

\section{Conclusion}
In this work, we introduced \dataset{}---a new dataset  that  focuses on commonsense-based implicit 
explanation
extraction from dialogues. The dataset consists of more than 4,500 manually annotated 
triplets from over 800 dialogues. We also introduced dialogue-level NLI and QA tasks, along with pre-trained transformer-based baselines to evaluate their inference and reasoning capabilities.
\section*{Acknowledgements}
This  research  is  supported  by A*STAR  under  its  RIE  2020  AME programmatic  grant,  Award No.–  A19E2b0098.
\bibliography{refs}
\bibliographystyle{acl_natbib}
\clearpage

\appendix

\section{Adversarial Filtering.}
For Task 3: Multi-choice Span Selection, once we have the collection $\mathcal{N}$, we follow \citet{zellers2018swag} to filter the confounding options generated in an iterative fashion. We follow the procedure below: 
\begin{enumerate}[wide, labelwidth=!, labelindent=0pt, itemsep=0pt]
    \item Initially, We select 3 random candidates from $\mathcal{N}$ and the correct answer to form a fake dataset.
    \item We split our fake dataset randomly into train and test set following a 1:2 ratio.
    \item We used our discriminator $D$ to filter out confounding options with unwanted stylistic patterns. Then, we train our discriminator $D$ on the dummy train set and score each option with a probability in the dummy test set.
    \item We replace the easiest confounding option (lowest probability) with another option from $\mathcal{N}$.
    \item We merge our dummy train set and dummy test set after replacement together to form our fake dataset for the next iteration
    \item Step 2,3,4,5 is repeated until the discriminator's cross-entropy loss converges.
\end{enumerate}  

We designed the input feed to $D$ as a combination of context $C$ and relation $R$, specifically we feed \code{<CLS> Conversation <SEP> Relation <SEP> Option\_i <SEP>} as input. Here \code{Option\_i} means the $i$th option in options. The probability score is given on the final layer vector corresponding to the \code{<CLS>} token. \\ 
We posit by excluding $A$ in our model input; the model can only pick up on low-level stylistic patterns with respect to the relation $R$ and context $C$ while not possessing reasoning abilities. Therefore, Our model can filter solely leveraging on low-level patterns while not based on the high-level inference. 
We use \code{roberta-base} model to filter out stylistic patterns. During the filtering process, discriminator prediction accuracy decreased from 0.55 to 0.27, suggesting the method's effectiveness in removing easy confounding candidates with stylistic patterns. 

\section{Additional Task}
\subsection{Task 4: Relation Prediction}
\label{sec:rp}
The fourth task of our interest is Relation Prediction between two spans from a conversation. 
Given two spans $A$ and $B$ from a conversation $C$, the task is to predict the unknown relation $R$ between them in $A \xrightarrow{?} B$.

We propose two different settings to evaluate the relation prediction task: 1)~Without Conversational Context and 2)~With Conversational Context.

\subsubsection{Task Description}
\paragraph{Without Conversational Context.} This setting is similar to the standard relation prediction task in knowledge graphs. Given the input spans $(A, B)$, the task is to predict the relation $R$ between $A$ and $B$.

\paragraph{With Conversational Context.} We surmise that the conversational context from $C$ is key to predict relation between any two given spans. This task setting is thus designed to evaluate that hypothesis. In this case, given the input spans and the conversation --- $(A, B, C)$, the task is to predict the commonsense relation $R$ between $A$ and $B$.

\subsubsection{Models}
We use pretrained transformer based models to benchmark this task as well. In particular, we used the \code{bert-base} and the \code{roberta-base} models. The input for the models is formulated as follows --- \code{<CLS> A <SEP> B <SEP>} in the without conversational context setting, and \code{<CLS> A <SEP> B <SEP> C <SEP>} in the with conversational context setting. The relation category $R$ is classified from the final layer vector corresponding to the \code{<CLS>} token.

\begin{table}[ht!]
\centering
\scalebox{0.66}{
\begin{tabular}{lll|ccccc|c}
\toprule
 &  & \textbf{Metric} & \textbf{Fold1} & \textbf{Fold2} & \textbf{Fold3} & \textbf{Fold4} & \textbf{Fold5} & \textbf{Avg.} \\ \midrule
\multirow{4}{*}{\rotatebox{90}{\small{BERT}}} & \multirow{4}{*}{\rotatebox{90}{\small{W/ Context}}} & Accuracy & 35.37 & 34.70 & 36.33 & 37.43 & 35.13 & 35.79 \\
 &  & Precision & 20.2 & 18.68 & 15.99 & 16.43 & 15.16 & 17.29 \\
 &  & Recall & 17.01 & 19.30 & 16.58 & 16.33 & 16.14 & 17.07 \\
 &  & F1 & 16.93 & 18.2 & 15.73 & 16.03 & 15.15 & 16.41 \\ \midrule
\multirow{8}{*}{\rotatebox{90}{\small{RoBERTa}}} & \multirow{4}{*}{\rotatebox{90}{\small{W/ Context}}} & Accuracy & 49.55 & 51.60 & 49.09 & 53.01 & 48.11 & 50.27 \\
 &  & Precision & 24.1 & 29.88 & 24.51 & 26.42 & 25.34 & 26.05 \\
 &  & Recall & 26.71 & 31.32 & 29.51 & 25.21 & 28.43 & 28.24 \\
 &  & F1 & 24.44 & 29.91 & 25.64 & 25.41 & 25.49 & 26.18 \\ \cline{2-9} 
 & \multirow{4}{*}{\rotatebox{90}{\small{W/O Context}}} & Accuracy & 39.46 & 41.32 & 36.33 & 40.39 & 39.49 & 39.40 \\
 &  & Precision & 17.00 & 19.72 & 14.44 & 16.77 & 15.99 & 16.78 \\
 &  & Recall & 18.51 & 17.22 & 15.90 & 14.36 & 16.27 & 16.45 \\
 &  & F1 & 16.29 & 18.26 & 13.52 & 15.28 & 14.77 & 15.62 \\ \bottomrule
\end{tabular}
}
\caption{Results for Task 2: Relation Prediction. All precision, recall and F1 scores are macro level measures.}
\label{tab:rp}
\end{table}
\subsubsection{Results}
The results for the relation prediction task is shown in \cref{tab:rp}. We report accuracy and other macro level scores in \cref{tab:rp}. We observe that the macro level scores are quite sub-par partly due to the fact that we have a lot of relations in the annotated dataset. It is also to be noticed that the incorporation of context brings a large improvement across all the evaluation metrics. The results support our hypothesis that contextual information is substantially important in predicting the relation between spans.

\section{Hyperparameters}
We use the AdamW~\cite{loshchilov2018decoupled} optimizer to train the models for all the tasks. More details about learning rate, batch size and epochs are given below.

\subsection{Hyperparameters for Task 1: NLI}
The \code{roberta-large-mnli} model is trained with a learning rate of $1e^{-5}$ and batch size of $8$ for $10$ epochs. 

\subsection{Hyperparameters for Task 2: Span Extraction}
The \code{roberta-base} and \code{span-bert} model are both trained with a learning rate of $1e^{-5}$ and batch size of $16$ for $12$ epochs. 

\subsection{Hyperparameters for Task 3: Multi-choice Span Selection}
\paragraph{Generating Confounding-Option Candidates.}
We used SpanBERT fine-tuned on SQUAD2.0 dataset, we trained using learning rate of $1e^{-5}$ and batch size of $16$ for $20$ epochs. 

\paragraph{Adversarial Filtering.}
We split dummy train and test portion randomly by using $2/3rd$ of dataset as train and $1/3rd$ of dataset as test. Every iteration, we only replace the option with lowest output score with other candidates. We continued for around 35 iteration before the loss converges.
We fine-tuned \code{roberta-base} model with learning rate of $5e^{-5}$, batch size of $16$ and $3$ epochs.

\paragraph{Answer Prediction.}
In the C\&Q set up, We trained \code{bert-base} and \code{roberta-base} with learning rate of $5e^{-5}$, $1e^{-5}$; batch size of $16$ and $48$ for $10$ and $20$ epochs respectively. In the Q set up, we used learning rate of $5e^{-5}$ and $1e^{-5}$ respectively with batch size of $16$ for $3$ epochs.

\subsection{Hyperparameters for Task 4: Relation Prediction}
For both \code{bert-base} and \code{roberta-base}, we used learning rate of $2e^{-5}$ and batch size of $32$ for $40$ epochs. 

\section{Relation Count}
The frequency of the categorized relations in the final annotated dataset is shown in \cref{tab:rcount}. Triplets having relation belonging to the \textit{Others} category were not considered in any of our four experiments.
\label{sec:relation_count}
\begin{table}[t]
  \centering
 \small
 \resizebox{0.75\linewidth}{!}{
  
   \begin{tabular}{l|l|r|r}
   \toprule
   Category & Relation & Instances & Category Total\\
   \midrule
   \multirow{9}{*}{Attribution} & Capable Of & 20 & \multirow{9}{*}{728} \\ 
   & Depends On & 9 & \\ 
   & Has A & 41 & \\ 
   & Has Property & 284 & \\
   & Has Subevent & 58 \\
   & Is A & 227 & \\
   & Manner Of & 60 & \\
   & NotHasProperty & 21 & \\
   & NotIsA & 8 & \\

   \midrule
   
   \multirow{6}{*}{Causal} & Causes & 1126 & \multirow{6}{*}{1958}\\ 
   & Causes Desire & 454 & \\ 
   & Implies & 338 & \\ 
   & NotCauses & 24 & \\
   & NotCauseDesire & 7 & \\
   & NotImplies & 9 & \\

   \midrule
   
   \multirow{4}{*}{Comparison} & Antonym & 25 & \multirow{4}{*}{98}\\ 
   & Distinct From & 20 & \\ 
   & Similar To & 30 & \\ 
   & Synonym & 23 & \\ 

   \midrule
   
   \multirow{1}{*}{Conditional} &  Has Prerequisite & 298 & \multirow{1}{*}{298}\\

   \midrule
   
   \multirow{5}{*}{Intentional} & Desires & 17 & \multirow{5}{*}{799}\\ 
   & Motivated By Goal & 361 & \\
   & Obstructed By & 244 & \\
   & Used For & 170 & \\
   & NotMotivatedByGoal & 7 & \\

   \midrule

   \multirow{1}{*}{Social} & Social Rule & 76 & \multirow{1}{*}{76}\\ 

   \midrule
   
   \multirow{2}{*}{Spatial} & At Location & 187 & \multirow{2}{*}{192}\\
   & Located Near & 5 & \\ 

   \midrule
   
   \multirow{3}{*}{Temporal} & Before & 119 & \multirow{3}{*}{237}\\ 
   & Happens On & 101 & \\ 
   & Simultaneous & 17 & \\ 

   \midrule

   \textit{Others} & \textit{Various} & 153 & \multirow{1}{*}{153}\\
   \midrule
   
   \multicolumn{2}{l|}{\textbf{Total}} & \textbf{4539} & \textbf{4539}\\
   
   \bottomrule
   \end{tabular}
   }
  \caption{\footnotesize Frequency of annotated relations in the dataset. The \textit{Others} category contains various relations such as \textit{Related To, Has Context} and several negated relations with very less frequency.}
  \label{tab:rcount}
\end{table}

\section{Question Template}
\begin{table}[ht!]
  \centering
 \resizebox{0.8\linewidth}{!}{
   \begin{tabular}{l|l|p{6cm}}
   \toprule
   Category & Relation & Question \\
   \midrule
   \multirow{7}{*}{Attribution} & Capable Of & What is \textbf{X} capable of? \\
   & Depends On & What does \textbf{X} depend on? \\
   & Has A & What does \textbf{X} have? \\
   & Has Property & What property does \textbf{X} have? \\
   & Has Subevent & What subevent does \textbf{X} have? \\
   & Is A & What is \textbf{X}? \\
   & Manner Of & What is \textbf{X} a manner of? \\
   \midrule
   
   \multirow{3}{*}{Causal} & Causes & What does \textbf{X} cause? \\
   & Causes Desire & What desire is caused by \textbf{X}? \\
   & Implies & What is implied by \textbf{X}? \\
   \midrule
   
   \multirow{4}{*}{Comparison} & Antonym & What is an antonym of \textbf{X}? \\
   & Distinct From & What is \textbf{X} distinct from? \\
   & Similar To & What is \textbf{X} similar to? \\
   & Synonym & What is a synonym of \textbf{X}? \\
   \midrule
   
   \multirow{1}{*}{Conditional} &  Has Prerequisite & What prerequisite does \textbf{X} have? \\
   \midrule
   
   \multirow{4}{*}{Intentional} & Desires & What does \textbf{X} desire? \\
   & Motivated By Goal & Which goal motivates the act/action \textbf{X}? \\
   & Obstructed By & What is \textbf{X} obstructed by? \\
   & Used For & What is \textbf{X} used for? \\
   \midrule

   \multirow{1}{*}{Social} & Social Rule & What is \textbf{X} the social norm for? \\
   \midrule
   
   \multirow{2}{*}{Spatial} & At Location & Where is \textbf{X} located? \\
   & Located Near & What is \textbf{X} located near?  \\
   \midrule
   
   \multirow{3}{*}{Temporal} & Before & What happens after \textbf{X}? \\
   & Happens On & When does \textbf{X} happen? \\
   & Simultaneous & What does \textbf{X} cooccur with? \\
   
   \bottomrule
   \end{tabular}
   }
  \caption{\footnotesize Question template in Task 2 and 3 for the various relations; \textbf{X} is the placeholder for head span $A$.}
  \label{tab:template}
\end{table}
\noindent Question templates used in Task 2: Span Extraction and Task 3: Multi-choice Span Selection is shown in \cref{tab:template}. The placeholder \textbf{X} in the Question is replaced with the actual annotated span $A$.

\end{document}